\documentclass[runningheads]{llncs}
\usepackage{graphicx}
\usepackage{amsmath,amssymb}
\def\bbeta{{\boldsymbol{\beta}}}
\begin{document}
\title{Evidence fusion with contextual discounting for multi-modality medical image segmentation \thanks{This work was supported by the China Scholarship Council (No. 201808331005). It was carried out in the framework of the Labex MS2T, which was funded by the French Government, through the program ``Investments for the future'' managed by the National Agency for Research (Reference ANR-11-IDEX-0004-02)}}
\titlerunning{Evidence fusion with contextual discounting}
\author{Ling Huang\inst{1,4} \and
Thierry Denoeux\inst{1,2} \and Pierre Vera \inst{3} \and 
Su Ruan\inst{4}}
\authorrunning{L. Huang et al.}
\institute{Heudiasyc, CNRS, Universit\'e de technologie de Compi\`egne, France \\ \email{ling.huang@hds.utc.fr} 
\and
Institut universitaire de France, France \and Department of Nuclear Medicine, Henri Becquerel Cancer Center, France \and Quantif, LITIS, University of Rouen Normandy, France }
\maketitle              % typeset the header of the contribution
\begin{abstract}
As information sources are usually imperfect, it is necessary to take into account their reliability in multi-source information fusion tasks. In this paper, we propose a new deep framework allowing us to merge multi-MR image segmentation results using the formalism of Dempster-Shafer theory while taking into account the reliability of different modalities relative to different classes. The framework is composed of an encoder-decoder feature extraction module, an evidential segmentation module that computes a belief function at each voxel for each modality, and a multi-modality evidence fusion  module, which assigns a vector of discount rates to each modality evidence and combines the discounted evidence using Dempster’s rule. The whole framework is trained by minimizing a new loss function based on a discounted Dice index to increase segmentation accuracy and reliability. The method was evaluated on the BraTs 2021 database of 1251 patients with brain tumors. Quantitative and qualitative results show that our method outperforms the state of the art, and implements an effective new idea for merging multi-information within deep neural networks.

\end{abstract}
\section{Introduction}
\label{sec:intro}
Single-modality medical images often do not contain enough information to reach a reliable diagnosis. This is why physicians generally use multiple sources of information, such as multi-MR images for brain tumor segmentation, or PET-CT images for lymphoma segmentation. The effective fusion of multi-modality information is of great importance in the medical domain for better disease diagnosis and radiotherapy. Using convolutional neural networks (CNNs), researchers have mainly adopted probabilistic approaches to information fusion, which can be classified into three strategies: image-level fusion such as input data concatenation \cite{peiris2021volumetric}, feature-level fusion such as attention mechanism concatenation \cite{zhou2020fusion}, and decision-level fusion such as weighted averaging \cite{kamnitsas2017ensembles}. However, probabilistic fusion is unable to effectively manage conflict that occurs when different labels are assigned to the same voxel based on different modalities. Also, it is important, when combining  multi-modality information, to take into account the reliability of the different sources \cite{mercier2008refined}. 

Dempster-Shafer theory (DST) \cite{dempster1967upper}\cite{shafer1976mathematical}, also known as belief function theory or evidence theory, is a formal framework for information modeling, evidence combination, and decision-making with uncertain or imprecise information \cite{denoeux20b}. Researchers from the medical image community have actively investigated the use of DST for handling uncertain, imprecision sources of information in different medical tasks, such as disease classification \cite{huang2021covid}, tumor segmentation \cite{huang2021evidential}\cite{huang2022lymphoma}, multi-modality image fusion \cite{huang2021deep}, etc. In the DST framework, the reliability of a source of information can be taken into account using the discounting operation \cite{shafer1976mathematical}, which transforms each piece of evidence provided by a source into a weaker, less informative one. 

In this paper, we propose a multi-modality evidence fusion framework with contextual discounting based on DST and deep learning. To our knowledge, this work is the first attempt to apply evidence theory with contextual discounting to the fusion of deep neural networks. The idea of considering multi-modality images as independent inputs and quantifying their reliability is simple and reasonable. However, modeling the reliability of sources is important and challenging. Our model computes mass functions assigning degrees of belief to each class and an ignorance degree to the whole set of classes. It thus has one more degree of freedom than a probabilistic model, which allows us to model source uncertainty and reliability directly. The contributions of this paper are the following: (1) Four DST-based evidential segmentation modules are used to compute the belief of each voxel belonging to the tumor for four modality MRI images; (2) An evidence-discounting mechanism is applied to each of the single-modality MRI images to take into account its reliability; (3) A multi-modality evidence fusion strategy is then applied to combine the discounted evidence with DST and achieve more reliable results. End-to-end learning is performed by minimizing a new loss function based on the discounting mechanism, allowing us to increase the segmentation performance and  reliability.

The overall framework will first be described in Section 2 and experimental results will be reported in Section 3.

\section{Methods}
\begin{figure}
\includegraphics[width=\textwidth]{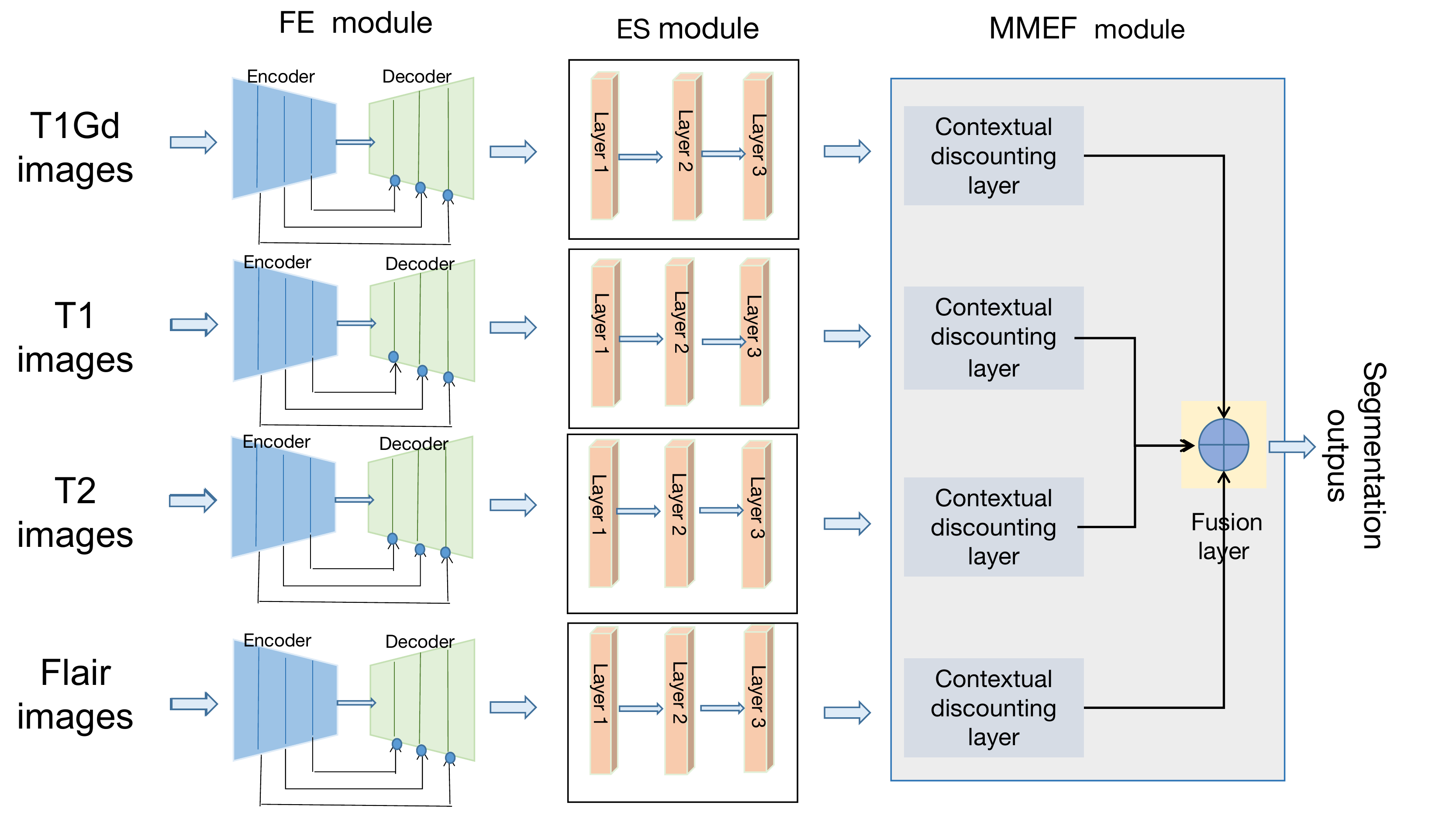}
\caption{Multi-modality evidence fusion framework. It is composed of four encoder-decoder feature extraction (FE) modules corresponding to T1Gd, T1, T2 and Flair modality inputs; four evidential segmentation (ES) modules corresponding to each of the inputs; and a multi-modality evidence fusion (MMEF) module.}
\label{fig:architecture}
\end{figure}

%\label{arch}
Fig.~\ref{fig:architecture} summarizes the proposed evidence fusion framework with contextual discounting. It is composed of (1) four encoder-decoder feature extraction (FE) modules (Residual-UNet \cite{kerfoot2018left}), (2) four evidential segmentation (ES) modules, and (3) a multi-modality evidence fusion (MMEF) module. Details about the ES and MMEF modules will be given in Sections \ref{subsec:ES} and  \ref{subsec:MMEF}, respectively. The discounted loss function used for optimizing the framework parameters will be described in Section \ref{subsec:loss}.

\subsection{Evidential Segmentation}
\label{subsec:ES}

\subsubsection{Dempster-Shafer theory.} Before presenting the evidential segmentation module, we first introduce some basic concepts of DST. Let $\Omega =\{\omega _{1}, \omega _{2}, \ldots, \omega _{K}\} $ be a finite set of hypotheses about some question. Evidence about $\Omega$ can be represented by a mapping $m$ from $2^\Omega$ to $[0,1]$ such that $\sum _{A\subseteq \Omega }m(A)=1$, called a \emph{mass function}. For any hypothesis $A\subseteq\Omega$, the quantity $m(A)$ represents the mass of belief allocated to $A$ and to no more specific proposition. The associated \emph{belief and plausibility functions} are defined, respectively, as 
\begin{equation}
  Bel (A)= \sum_{\emptyset \ne B\subseteq A} m(B) \quad \text{and} \quad   Pl(A)=\sum_{  B\cap  A \ne \emptyset} m(B), \quad \forall A \subseteq \Omega.
  \label{eq:bel}
\end{equation}
The \emph{contour function} $pl$ associated to $m$ is the function that maps each element $\omega$ of $\Omega$ to its plausibility: $pl(\omega)=Pl(\{\omega\})$.

Two mass functions $m_{1}$ and $m_{2}$ derived from two independent items of evidence can be combined by \emph{Dempster's rule} \cite{shafer1976mathematical}, denoted as $\oplus$ and formally defined by $(m_{1}\oplus m_{2})(\emptyset)=0$ and
\begin{equation}
    (m_{1}\oplus m_{2})(A)=\frac{1}{1-\kappa }\sum _{B\cap C=A}m_{1}(B)m_{2}(C),
    \label{eq:demp}
\end{equation}
for all $A\subseteq \Omega, A\neq \emptyset$, where $\kappa$ represents the \emph{degree of conflict} between $m_{1}$ and $m_{2}$: 
\begin{equation}
    \kappa=\sum _{B\cap C=\emptyset}m_{1}(B)m_{2}(C).
    \label{eq:conflict}
\end{equation}
The combined mass $m_{1}\oplus m_{2}$ is called the \emph{orthogonal sum} of $m_1$ and $m_2$. Let $pl_1$, $pl_2$ and $pl_{12}$ denote the contour functions associated with, respectively, $m_1$, $m_2$ and $m_{1}\oplus m_{2}$. the following equation holds:
\begin{equation}
\label{eq:prodpl}
    pl_{12}=\frac{pl_1 pl_2}{1-\kappa}.
\end{equation}
Equation \eqref{eq:prodpl} allows us to compute the contour function in time proportional to the size of $\Omega$, without having to compute the combined mass $m_{1}\oplus m_{2}$.

\subsubsection{Evidential segmentation module}
Based on the evidential neural network model introduced in \cite{denoeux2000neural} and using an approach similar to that recently described in \cite{huang2021evidential}, we propose a DST-based evidential segmentation module to quantify the uncertainty about the class of each voxel by a mass function. The basic idea of the ES module is to assign a mass to each of the $K$ classes and to the whole set of classes $\Omega$,  based on the distance between the feature vector of each voxel and $I$ prototype centers. The input feature vector can be obtained using any FE module, e.g., Residual-UNet \cite{kerfoot2018left}, nnUNet \cite{isensee2018nnu}.

The ES module is composed of an input layer, two hidden layers and an output layer. The first input layer is composed of $I$ units, whose weights vectors are prototypes $p_1,\ldots, p_I$ in input space. The activation of unit $i$ in the prototype layer for input vector $x$ is
\begin{equation}
    s_i=\alpha _i \exp(-\gamma_i \| x-p_i\|^2),   
    \label{eq:enn1}
\end{equation}
where $\gamma_i>0$ and $\alpha_i \in [0,1]$ are two parameters. The second hidden layer computes mass functions $m_i$ representing the evidence of each prototype $p_i$ using the following equations:
\begin{subequations}
\begin{align}
m_i(\{\omega _{k}\})&=u_{ik}s_i, \quad k=1,..., K\\
m_{i}(\Omega)&=1-s_i, 
\label{eq:enn2}
\end{align}
\end{subequations}
where $u_{ik}$ is the membership degree of prototype $i$ to class $\omega_k$, and $\sum _{k=1}^K u_{ik}=1$. Finally, the third layer combines the $I$ mass functions $m_1,\ldots,m_I$ using Dempster's combination rule \eqref{eq:demp} to finally obtain a belief function quantifying uncertainty about the class of each voxel.

\subsection{Multi-modality evidence fusion}
\label{subsec:MMEF}
In this paper, the problem of quantifying the source reliability is addressed by the discounting operation of DST. Let $m$ be a mass function on $\Omega$ and $\beta$ be a coefficient in $[0,1]$. The \emph{discounting} operation \cite{shafer1976mathematical} with discount rate $1-\beta$ transforms $m$ into a weaker, less informative mass function $^\beta m$ defined as 
\begin{equation}
\label{eq:disc}
^\beta m=\beta \, m +(1-\beta) \,m_?,
\end{equation}
where $m_?$ is the vacuous mass function defined by $m_?(\Omega)=1$, and coefficient $\beta$ is the degree of belief that the source mass function $m$ is reliable \cite{smets94a}. When $\beta=1$ we accept mass function $m$ provided by the source and take it as a description of our knowledge; when $\beta=0$, we reject it and we are left with the vacuous mass function $m_?$. In this paper, we focus on the situation when $\beta \in [0,1]$ and combine uncertain evidence with partial reliability using Dempster's rule.

As suggested in \cite{mercier2008refined}, the above discounting operation can be extended to \emph{contextual discounting}. This operation can represent richer meta-knowledge regarding the reliability of the source of information in different contexts \cite{denoeux19f}. It is parameterized by a vector $\bbeta=(\beta_1,\ldots,\beta_K)$, where $\beta_k$ is  the degree of belief that the source is reliable given that the true class is $\omega_k$. The complete expression of the discounted lass function is given in \cite{mercier2008refined}. Here, we just give the expression of the  corresponding contour function, which will be used later:
\begin{equation}
     ^\bbeta pl(\{\omega_k \})=1-\beta_k+\beta_k  pl(\{\omega_k\}), \quad k=1,\ldots,K,
     \label{eq:c_discounting}
\end{equation}
When we have several sources provided by independent evidence, the discounted evidence can be combined by Dempster's rule. Assuming that we have two source of information, let $^{\bbeta_1} pl_{S_{1}}$ and $^{\bbeta_2} pl_{S_{2}}$ be the discounted contour functions provided, respectively, by sources $S_1$ and $S_2$, with discount rate vectors $1-\bbeta_1$ and $1-\bbeta_2$. From \eqref{eq:prodpl}, the combined contour function is proportional to the product $ {^{\bbeta_1}} pl_{S_1} {^{\bbeta_2}} pl_{S_2}$.  

\subsection{Discounted Dice loss}
\label{subsec:loss}
We define a new loss function based on discount rates, hereafter referred to as the discounted Dice loss, given as follows:
\begin{subequations}
\begin{equation}
    \textsf{loss}_{D}=1-\frac{2 \sum_{n=1}^{N} {^\beta} S_n  G_n}{ \sum_{n=1}^{N} {^\beta} S_n + \sum_{n=1}^{N} G_n},
\label{eq:loss}
\end{equation}  
where $ ^\beta S_n$ is the normalized segmentation output for voxel $n$ by fusing $H$ discounted source information,
\begin{equation}
    ^\beta S_n= \frac {\prod_{h=1}^{H} {^{\beta^h}} pl_{S_h}(\{ \omega_k \})} {\sum_{k=1}^{K} \prod_{h=1}^{H} {^{\beta^h}} pl_{S_h}(\{ \omega_k \})},
 \end{equation}
\end{subequations}
 $G_n$ is the ground truth for voxel $n$, and $N$ is the number of voxels in the volume.

\section{Experimental results}
In this section, we report results of our methods on the brain tumor segmentation dataset. The experimental settings will be described in Section \ref{subsec:settings}. A comparison with the state of the art and an ablation study will be reported in Section \ref{subsec:comp}.

\subsection{Experiment settings}
\label{subsec:settings}
\subsubsection{BraTS Dataset} We used 1251 MRI scans with size  $240\times 240 \times 155$ voxels coming from the BraTS 2021 Challenge \cite{baid2021rsna} \cite{menze2014multimodal} \cite{bakas2017advancing} to evaluate our framework. For each scan, there are four modalities: native (T1), post-contrast T1-weighted (T1Gd), T2-weighted (T2), and T2 Fluid Attenuated Inversion Recovery (FLAIR). Annotations of 1251 scans comprise the GD-enhancing tumor (ET-label 4), the peritumoral edema (ED-label 2), and the necrotic and non-enhancing tumor core (NCR/NET-label1). The task is to segment three semantically meaningful tumor classes: the Enhancing Tumor (ET), the Tumor Core (TC) region (ET+NCR/NET), and the Whole Tumor (WT) region (ET+NCR/NET+ED). Following \cite{peiris2021volumetric}, we divided the 1251 scans into 834, 208, 209 scans for training, validation and testing, respectively. Using the same pre-processing operation as in \cite{peiris2021volumetric}, we performed min-max scaling operation followed by clipping intensity values to standardize all volumes, and cropped/padded the volumes to a fixed size of $128 \times 128 \times 128$ by removing unnecessary background. %No data augmentation technique was applied and no additional data was used in this study.

\subsubsection{Implementation details}
The initial values of parameters $\alpha_i$ and $\gamma_i$ were set, respectively, to 0.5 and 0.01, and membership degrees $u_{ik}$ were initialized randomly by drawing uniform random numbers and normalizing. The prototypes were initialized by the $k$-means clustering algorithm. Details about the initialization of ES module parameters can be found in \cite{huang2022lymphoma}. For the MMEF module, the initial values of parameter $\beta_k$ was set to 0.5. Each model was trained on the learning set with 100 epochs using the Adam optimization algorithm. The initial learning rate was set to $10^{-3}$. %An adjusted learning rate schedule was applied by reducing the learning rate when the training loss did not decrease in 10 epoch. 
The model with the best performance on the validation set was saved as the final model for testing\footnote{Our code will be available at \url{https://github.com/iWeisskohl}.}.  %The framework was implemented in Python with the PyTorch-based medical image framework MONAI, and was trained and tested on a desktop with a 2.20GHz Intel(R) Xeon(R) CPU E5-2698 v4 and a Tesla V100-SXM2 graphics card with 32 GB GPU memory.

\subsection{Segmentation Results}
\label{subsec:comp}

\begin{table}%[ht]
  \centering
  \caption{Segmentation Results on the BraTS 2021 dataset. The best result is shown in bold, and the second best is underlined.}
  \label{sota}
  \begin{tabular}{llllllllllllllll}
  \hline
  \multicolumn{1}{l}{Methods}  &\multicolumn{7}{c}{Dice score} &\multicolumn{7}{c}{Hausdorff distance}\\
  \cline{2-8}
  \cline{10-16}
 & ET&&TC&&WT&&Avg && ET&&TC&&WT&&Avg\\
\hline
UNet \cite{cciccek20163d}&83.39& &86.28 &&89.59&& 86.42& &11.49&& 6.18&& 6.15 &&7.94\\
VNet \cite{milletari2016v} & 81.04&& 84.71 &&90.32&& 85.36&& 17.20 &&7.48&&  7.53 &&10.73\\
nnFormer \cite{zhou2021nnformer}&82.83 &&86.48 &&90.37&& 86.56&&11.66 && 7.89 && 8.00 &&9.18\\
VT-UNet \cite{peiris2021volumetric}&85.59 &&87.41 &&91.20& &88.07&& \textbf{10.03} &&6.29&&  6.23 && \underline{7.52}\\

Residual-UNet \cite{kerfoot2018left} &	85.07 &&87.61 &&89.78&&87.48 && 11.76 && \underline{6.14}&& 6.31 && 8.07 	\\
nnUNet \cite{isensee2018nnu} &\underline{87.12}&& \textbf{90.31}&& \underline{91.47}&& \underline{89.68} && 12.46 && 11.04&& 5.97&&9.82\\
MMEF-UNet (Ours) & 86.96 && 87.46 && 90.68& &88.36&& \underline{10.20} && \textbf{6.07} &&  \underline{5.29} && \textbf{7.18}\\
MMEF-nnUNet (Ours) &\textbf{87.26} &&\underline{90.05} &&\textbf{92.83}& &\textbf{90.05} && 10.09&& 9.68&& \textbf{5.10} &&8.29 \\
\hline
\end{tabular}
\end{table}
\subsubsection{Segmentation Accuracy} We used the Dice Score and the Hausdorff Distance (HD) as our evaluation metrics. For each patient, we separately computed these two indices for three classes and then averaged indices over the patients, following a similar evaluation strategy as in \cite{peiris2021volumetric}. We used two baseline models, Residual-UNet \cite{kerfoot2018left} and nnUNet \cite{isensee2018nnu}, as FE modules and applied our proposal based on the two corresponding models to construct our methods named, respectively, MMEF-UNet and MMEF-nnUNet. We compared our methods with two recent transformer-based models (nnFormer \cite{zhou2021nnformer}, VT-UNet \cite{peiris2021volumetric}), two classical CNN-based methods (UNet \cite{cciccek20163d}, V-Net \cite{milletari2016v}), and the two baseline methods. The quantitative results are reported in Table~\ref{sota}. Our methods outperform the two classical CNN-based models and two recent transformer-based methods according to the Dice score, the best result being obtained by MMEF-nnUNet according to this criterion. In contrast, MMEF-UNet achieves the lowest HD. 

\subsubsection{Segmentation Reliability} To test the reliability of our methods, we computed the Expected Calibration Error (ECE) \cite{guo2017calibration} for the two baseline methods and our methods. We obtained ECE values of 2.35 \% and 2.04\%, respectively, for Residual-UNet and MMEF-UNet, against ECE values of 4.46\% and 4.05\%, respectively, for nnUNet and MMEF-nnUNet, respectively. The probabilities computed by our models thus appear to be better calibrated. 
\subsubsection{Ablation Study} 
We also investigated the contribution of each module component to the performance of the system. Table~\ref{ablition} highlights the importance of introducing the ES and MMEF modules. Residual-UNet is the baseline model that uses the softmax transformation to map feature vectors to probabilities. Compared to Residual-UNet, Residual-UNet-ES uses the ES module instead of softmax. Residual-UNet-ES-MMEF, our final proposal, fuses the four single modality outputs from Residual-UNet-ES with MMEF module. Compared to the baseline method Residual-UNet, our method, which plugs the ES module after Residual-UNet, improves the segmentation performance based on single T1Gd, T1 and Flair inputs. Furthermore, the use of the MMEF module improves the performance to a large amount compared to any single modality. 

\begin{table}%[ht]
  \centering
  \caption{Segmentation Results on BraTS 2021 Data ($\uparrow $ means higher is better).}
  \centering
  \label{ablition}
  \begin{tabular}{lllllllllll}
  \hline
  \multicolumn{1}{l}{Methods}& &\multicolumn{1}{l}{Input Modality}&  &\multicolumn{7}{c}{Dice}\\
  \cline{5-11}
   && &&ET&&ED&&NRC/NET &&Avg\\
\hline
Residual-UNet&&T1Gd&&84.13 $\uparrow $&&67.44&&72.64&&74.73\\
Residual-UNet-ES&&T1Gd&&83.92&&68.34 $\uparrow $&&73.29$\uparrow $&&75.18$\uparrow $\\
\hline
Residual-UNet&&T1&&53.67&&63.51&&42.81&&53.33\\
Residual-UNet-ES&&T1&&57.26$\uparrow $&&67.83$\uparrow $&&54.77$\uparrow $&&59.95$\uparrow $\\
\hline
Residual-UNet&&T2&&55.49$\uparrow $&&69.44&&51.16&&58.70$\uparrow $\\
Residual-UNet-ES&&T2&&54.96&&69.84$\uparrow $&&51.27$\uparrow $&&58.69\\
\hline
Residual-UNet&&Flair&& 50.48&&75.16&&38.06&&54.56\\
Residual-UNet-ES&&Flair&&53.67$\uparrow $&&77.22$\uparrow $&&49.71$\uparrow $&&60.20$\uparrow $\\
\hline
%EUnet+Fusion&T1Gd,T1,T2,Flair&\\
Residual-UNet-ES-MMEF&&T1Gd,T1,T2,Flair&&86.96$\uparrow $&&85.48$\uparrow $&&78.98$\uparrow $&&83.81$\uparrow $\\
\hline
\end{tabular}
\end{table}

\begin{table}%[ht]
  \centering
  \caption{Reliability value $\beta$ (after training) for classes ET, ED and NRC/NET and the four modalities. Higher values correspond to greater contribution to the segmentation.}
  \label{beta}
  \begin{tabular}{lcccc}
  \hline
  \multicolumn{1}{l}{$\beta$}& &\multicolumn{1}{l}{ET}  & \multicolumn{1}{l}{ED}&
  \multicolumn{1}{l}{NRC/NET} \\
  \hline
  T1Gd&&0.9996 &	0.9726  &0.9998\\
  T1&&0.4900&	0.0401&	0.2655\\
  T2&&0.4814&0.3881&	0.4563\\
  Flair&&0.0748&	0.86207	&0.07512\\
\hline
\end{tabular}
\end{table}

\subsubsection{Interpretation of reliability coefficients} 
\begin{figure}%[ht]
\centering
\includegraphics[width=\textwidth]{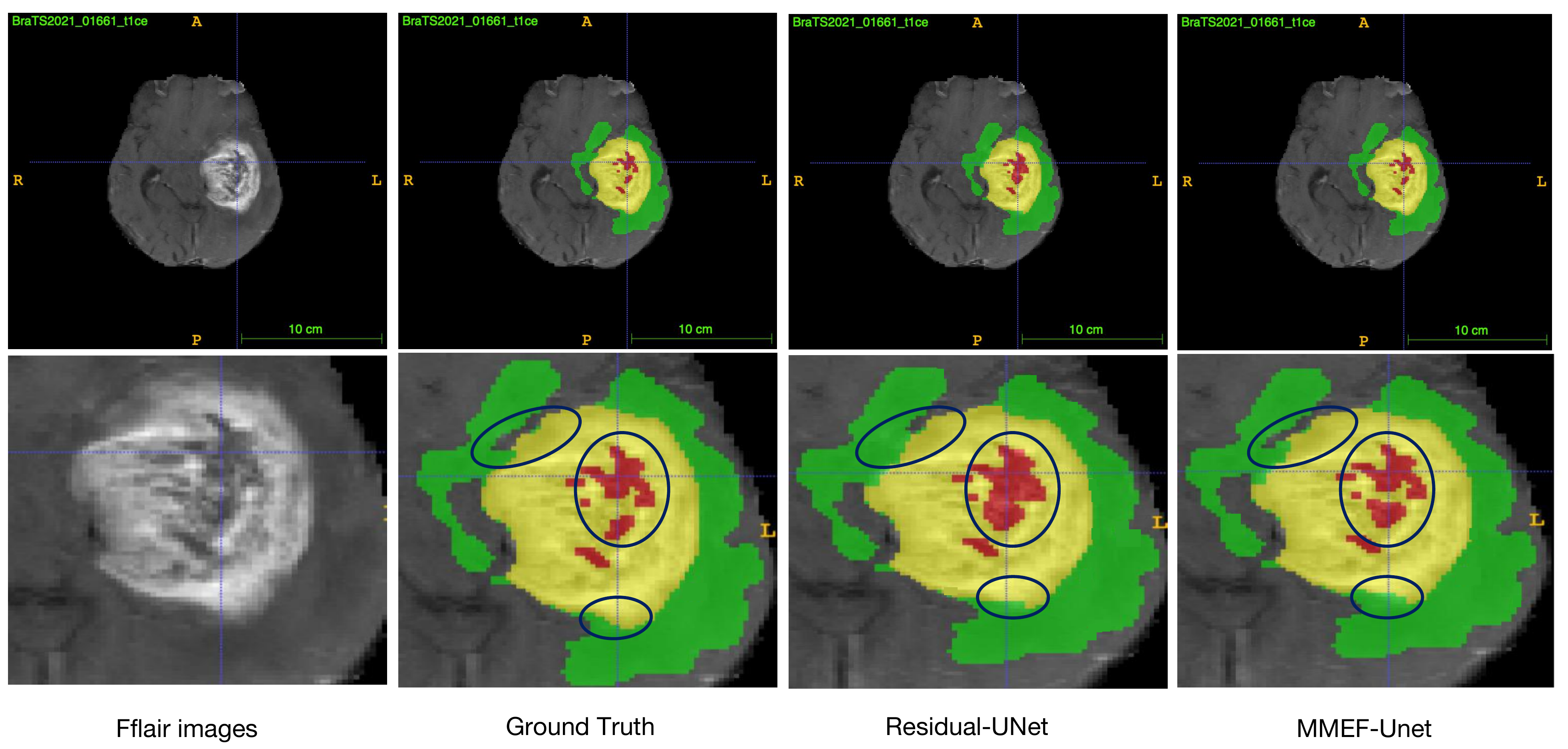}
\caption{Visualized segmentation results. The first and the second row are the whole brain with tumor and the detailed tumor region (the main differences are marked in blue circles). The three columns correspond, from left to right, to the Flair image, the ground truth and the segmentation results obtained by Residual-UNet and MMEF-UNet. The green, yellow and red represent the ET, ED and NRC/NET, respectively.}
\label{fig:visu}
\end{figure}

Table~\ref{beta} shows the learnt reliability values $\beta$ for the four modalities with three different classes. The evidence from the T1Gd modality is reliable for ET, ED and NRC classes with the highest $\beta$ value. In contrast, the evidence from the Flair modality is only reliable for the ED class, and the reliability value $\beta$ of the T2 modality is only around 0.4 for three classes. The evidence from the T1 modality is less reliable for the three classes compared to the evidence of the other three modalities. These reliability results are consistent with domain knowledge about these modalities reported in \cite{baid2021rsna}. Fig.~\ref{fig:visu} shows the segmentation results of Residual-UNet with the inputs of four concatenated modalities and MMEF-UNet with the inputs of four separate modalities. Our model  locates and segments brain tumors precisely, especially the ambiguous voxels located at the tumor boundary.

\section{Conclusion}
Based on DST, a multi-modality evidence fusion framework considering segmentation uncertainty and source reliability has been proposed for multi-MRI brain tumor segmentation. The ES module performs tumor segmentation and uncertainty quantification, and the MMEF module allows for multi-modality evidence fusion with contextual discounting and Dempster's rule. %Qualitative and quantitative evaluations show promising results compared to the baseline and state-of-the-art methods. 
This work is the first to implement contextual discounting for the fusion of multi-modal information with DST and DNN. The contextual discounting operation allows us to take into account the uncertainty of the different sources of information directly, and it reveals the reliability of different modalities in different contexts. Our method can be used together with any state-of-the-art FE module to get better performance. 

Some limitations of this work remain in the computation cost and the segmentation accuracy. We treat single modality images as independent inputs using  independent FE and ES modules, which introduces additional computation costs compared to image concatenation methods (e.g., the FLOPs and parameter numbers are equal to 280.07G and 76.85M for UNet-MMEF, against 73.32G, and 19.21M for Residual-UNet). In future research, we will refine our algorithm to improve the accuracy and reliability of our model, and reduce its complexity. We will also explore the possibility of cross-modality evidence fusion for survival prediction tasks.

 \bibliographystyle{splncs04}
 \bibliography{ref}
%Author, F.: Article title. Journal \textbf{2}(5), 99--110 (2016)

%Author, F., Author, S.: Title of a proceedings paper. In: Editor, F., Editor, S. (eds.) CONFERENCE 2016, LNCS, vol. 9999, pp. 1--13. Springer, Heidelberg (2016). \doi{10.10007/1234567890}

%Author, F., Author, S., Author, T.: Book title. 2nd edn. Publisher,Location (1999)

%Author, A.-B.: Contribution title. In: 9th International Proceedings on Proceedings, pp. 1--2. Publisher, Location (2010)

%LNCS Homepage, \url{http://www.springer.com/lncs}. Last accessed 4 Oct 2017
%\end{thebibliography}

\end{document}